\documentclass[a4paper]{article}

\usepackage{INTERSPEECH2015}

\usepackage{graphicx}
\usepackage{amssymb,amsmath,bm}
\usepackage{textcomp,array}
\usepackage[normalem]{ulem}

\usepackage{xspace}
\usepackage{url}
\usepackage{xargs} 

\newcommand{\word}[1]{\textsl{#1}}        
\newcommand \ignore[1]{}

\newcolumntype{P}[1]{>{\centering\arraybackslash}p{#1}}
\newcolumntype{M}[1]{>{\centering\arraybackslash}m{#1}}

\ninept

\title{DNN-based Speech Synthesis for Indian Languages from ASCII text}

\makeatletter
\def\name#1{\gdef\@name{#1\\}}
\makeatother \name{{\em Srikanth Ronanki$^1$, Siva Reddy$^2$, Bajibabu Bollepalli$^3$, Simon King$^1$}}

\address{$^1$The Centre for Speech Technology Research, University of Edinburgh, United Kingdom \\
  $^2$ILCC, School of Informatics, University of Edinburgh, United Kingdom \\
  $^3$Department of Signal Processing and Acoustics, Aalto University, Finland \\
  {\small \tt srikanth.ronanki@ed.ac.uk}
}

\begin{document}

  \maketitle
\begin{abstract}
Text-to-Speech synthesis in Indian languages has a seen lot of progress over the decade partly due to the annual Blizzard challenges. These systems assume the text to be written in Devanagari or Dravidian scripts which are nearly phonemic orthography scripts. However, the most common form of computer interaction among Indians is ASCII written transliterated text. Such text is generally noisy with many variations in spelling for the same word. In this paper we evaluate three approaches to synthesize speech from such noisy ASCII text: a naive Uni-Grapheme approach, a Multi-Grapheme approach, and a supervised Grapheme-to-Phoneme (G2P) approach. These methods first convert the ASCII text to a phonetic script, and then learn a Deep Neural Network to synthesize speech from that. We train and test our models on Blizzard Challenge datasets that were transliterated to ASCII using crowdsourcing. Our experiments on Hindi, Tamil and Telugu demonstrate that our models generate speech of competetive quality from ASCII text compared to the speech synthesized from the native scripts. All the accompanying transliterated datasets are released for public access.
\end{abstract}


\section{Introduction}
\label{sec:intro}

Though a large number of Indian languages have indigenous scripts, the lack of a standardized keyboard, and the ubiquity  of QWERTY keyboards, means that people most often write using ASCII\footnote{The ASCII character set is the union of Roman alphabets, digits, and a few punctuation marks.} \cite{ascii86} text using spellings motivated largely by pronunciation \cite{ahmed-EtAl:2011:WTIM2011}. Increasingly, many technologies such as Web search and natural language processing are adapting to this phenomenon \cite{Roy:2013:OFT:2701336.2701636,Gupta:2014:QEM:2600428.2609622,vyas-EtAl:2014:EMNLP2014}. In the area of Speech Synthesis, although the efforts of the 2013, 2014 and 2015 Blizzard Challenges\footnote{\url{http://www.synsig.org/index.php/Blizzard_Challenge}} \cite{prahallad2014blizzard,prahallad2013blizzard} resulted in improvements to the naturalness of speech synthesis of Indian languages, the text was assumed to be written in native script. In this work, we transliterate Blizzard data to informal chat-style ASCII text using Mechanical Turkers, and synthesize speech from the resulting transliterated ASCII text. This represents a more realistic use case than in the Blizzard Challenge. 

Synthesizing speech from ASCII text is challenging: Since there is no standard way to spell pronunciations, people often spell same word in multiple ways, e.g., the word \word{start} in Telugu can be ASCII spelled \word{prarhambham}, \word{prarambham}, \word{prarambam}, \word{praranbam}, etc. whilst words that differ in both pronunciation and meaning might be spelled the same, e.g., the words \word{ledhu} and \word{ledu} in Telugu could both be spelled \word{ledu}.

We address these problems by first converting ASCII graphemes to phonemes, followed by a DNN to synthesise the speech. We propose three methods for converting graphemes to phonemes. The first model is a naive model which assumes that every grapheme corresponds to a phoneme. In the second model, we enhance the naive model by treating frequently co-occurring character bi-grams as additional phonemes. In the final model, we learn a Grapheme-to-Phoneme transducer from parallel ASCII text and gold-standard phonetic transcriptions.

\ignore{Generation of synthetic speech in Indian languages from noisy ASCII text which to our knowledge is the first such application of G2P based speech synthesis.}

\noindent The contributions of this paper are:
\begin{itemize}
  \setlength\itemsep{0em}
 \item to synthesize speech from ASCII transliterated text for Indian languages, which to our knowledge is the first such attempt. Our results show that our Grapheme-to-Phoneme conversion model combined with a DNN acoustic model performs competitively with state-of-the-art speech synthesizers that use native script text.
\item the release of parallel ASCII transliterations of Blizzard data to foster research in this area.
\end{itemize}

\section{Related work}


\subsection{Transliteration of Indian Languages}

Many standard transliteration systems exist for Indian languages. Table~\ref{table1} shows different transliterations for an example sentence. Among these, CPS (Common Phone Set) \cite{bib7} and IT3 \cite{lavanya2005simple} are widely used by the speech technology community, ITRANS\footnote{\url{https://en.wikipedia.org/wiki/ITRANS}} \cite{madhavi2005om} is used in publishing houses, and WX\footnote{\url{https://en.wikipedia.org/wiki/WX_notation}} \cite{gupta2010transliteration} by the Natural Language Processing (NLP) community. Though these scripts provide umambiguous conversion to native Indian scripts, due to their lack of readability, and the overhead in learning how to use them, people still spell their words motivated by pronunciation. One such transliteration is shown in the row \word{Informal} of Table \ref{table1}. 

The most common trend observed in the literature is to treat transliteration as a machine translation and discriminative ranking problem \cite{Li:2009:RNM:1699705.1699707}. Our work aims to exploit the fact that transliterations are phonetically motivated, and therefore treat transliteration as a conversion problem. Specifically, we convert informal transliterations to  phonetic script, and then synthesize speech from the phonetic script using a DNN.

\begin{table}[t]
  \centering
\caption{Transliteration of Hindi text in various notations}
  \begin{tabular}{|M{2cm}|M{5.5cm}|}
    \hline
    Original Sentence & \vspace{-0.1em} \includegraphics{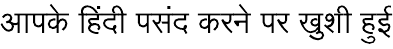}  \\ \hline \hline
    \bf{Notation} & \bf{Transliteration} \\ \hline
    CPS & aapakei hiqdii pasaqda karanei para khushii huii \\ \hline
    IT3 & aapakei hin:dii pasan:da karanei para khushii huii  \\ \hline
    ITRANS & Apake hiMdI pasaMda karane para khushI huI \\ \hline
    WX & Apake hiMdI pasaMda karane para KuSI huI \\ \hline
    Informal & apke hindi pasand karne par kushi hui \\ \hline
  \end{tabular}
\label{table1}
\end{table}

\begin{table}[t]
\centering
\caption{Training data for the Grapheme-to-Phoneme model}
\label{pronunciationTable}
\begin{tabular}{|M{1.4cm}|M{2cm}|M{3.5cm}|}
\hline
Word & Informal transliteration & Pronunciation (CPS notation) \\ \hline
 \begin{minipage}{1.1cm}
      \includegraphics[width=9mm]{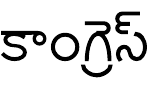}
    \end{minipage} & congress &   /k/aa/q/g/r/e/s/  \\ \hline
\begin{minipage}{1.3cm}
      \includegraphics[height=5mm,width=15mm]{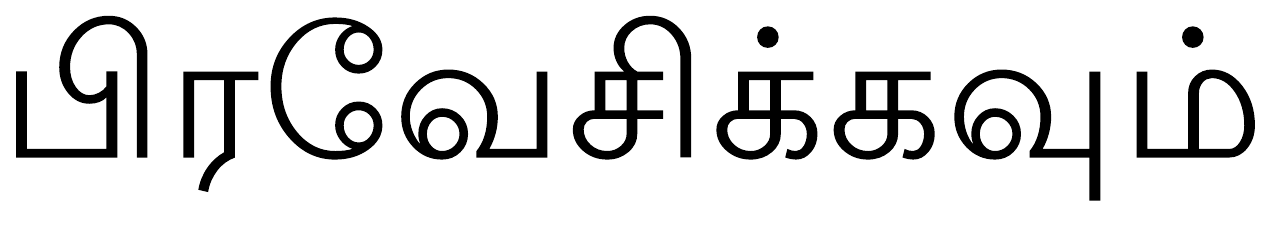}
    \end{minipage} & pravesikkavum & /p/i/r/a/w/ei/c/i/k/k/a/w/u/m/ \\ \hline
\begin{minipage}{1.1cm}
      \includegraphics[width=9mm]{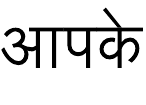}
    \end{minipage} & aapke & /aa/p/a/k/e/ \\ \hline
\end{tabular}
\end{table}

\subsection{Statistical Speech Synthesis}

Most existing work in speech synthesis for Indian languages uses unit selection \cite{bib9} with syllable-like units \cite{bib12,bib6}. Recently, based on the observation that Indian languages share many commonalities in phonetics, a language independent phone set was proposed, and was used in building statistical parametric (HMM-based) speech synthesis systems \cite{bib7}. We make use of this common phone set in one of our models. 

Our work also aligns with the recent literature on unsupervised learning for text-to-speech synthesis which aims to reduce the reliance on human knowledge and the manual effort required for building language-specific resources \cite{bib16,bib8,watts2015nst}. These approaches are able to learn from noisy input representations where there is no standard orthography. Following the success of DNNs for speech recognition \cite{bib3} and synthesis \cite{zen2013dnn, wu2015dnn, bib15}, we also use a DNN as the acoustic model.

\section{Our Approach}
\label{sec:approaches}
Our speech synthesis pipeline consists of two steps: 1)~Converting the input ASCII transliterated text to a phonetic script; 2)~learning a DNN based speech synthesizer from the parallel phonetic text and audio signal.

\subsection{Converting ASCII text to Phonetic Script}
\label{sec:approachesA}

We explore three different approaches which vary in the degree of supervision in defining a phoneme.

\subsubsection{Uni-Grapheme Model}
\label{ssec:Grapheme}

In this approach, we assume each ASCII grapheme acts as a phoneme. We assume that the DNN will learn to map these ``phonemes'' to speech sounds. We normalize the data to lowercase and remove all punctuation marks. This ensures that the phone-set contains 26 letters and an extra $/sil/$ phone to mark beginning and end of the sentence. 

\subsubsection{Multi-Grapheme Model}
\label{ssec:Hybrid}

In this approach, in addition to uni-graphemes, we also include some frequently co-occurring bi-graphemes as ``phonemes''. From manual inspection of the top 50 bi-graphemes, we found that the phonemes indicating stop consonants such as $/kh/$, $/ch/$, $/th/$, $/ph/$, $/bh/$ and long vowels such as $/aa/$, $/ii/$, $/ee/$, $/oo/$, $/uu/$ and dipthongs such as $/ai/$, $/au/$, $/ou/$ appear most frequently across languages. We selected 17~of these bi-graphemes as phonemes in addition to the above 27~uni-graphemes, making a total of 44~phonemes.

\subsubsection{Grapheme-to-Phoneme (G2P) Model}
\label{ssec:g2p}

In this model, we assume the phoneme-set is given. We use the common phone set (CPS, \cite{bib7}) to work with the languages of interest. We convert the native text to CPS~phonetic text using deterministic converters \cite{lavanya2005simple, raj2007text}. We then align the phonetic transcriptions to the ASCII transliterations from Mechanical Turkers to create a pronunciation table. Table \ref{pronunciationTable} shows the parallel data with the native text in the first column, the informal ASCII transliteration in the second column, and the CPS phonetic transcription in the third column. 

\begin{figure}[t]
\begin{minipage}[b]{1.0\linewidth}
  \centering
  \centerline{\includegraphics[width=8.5cm]{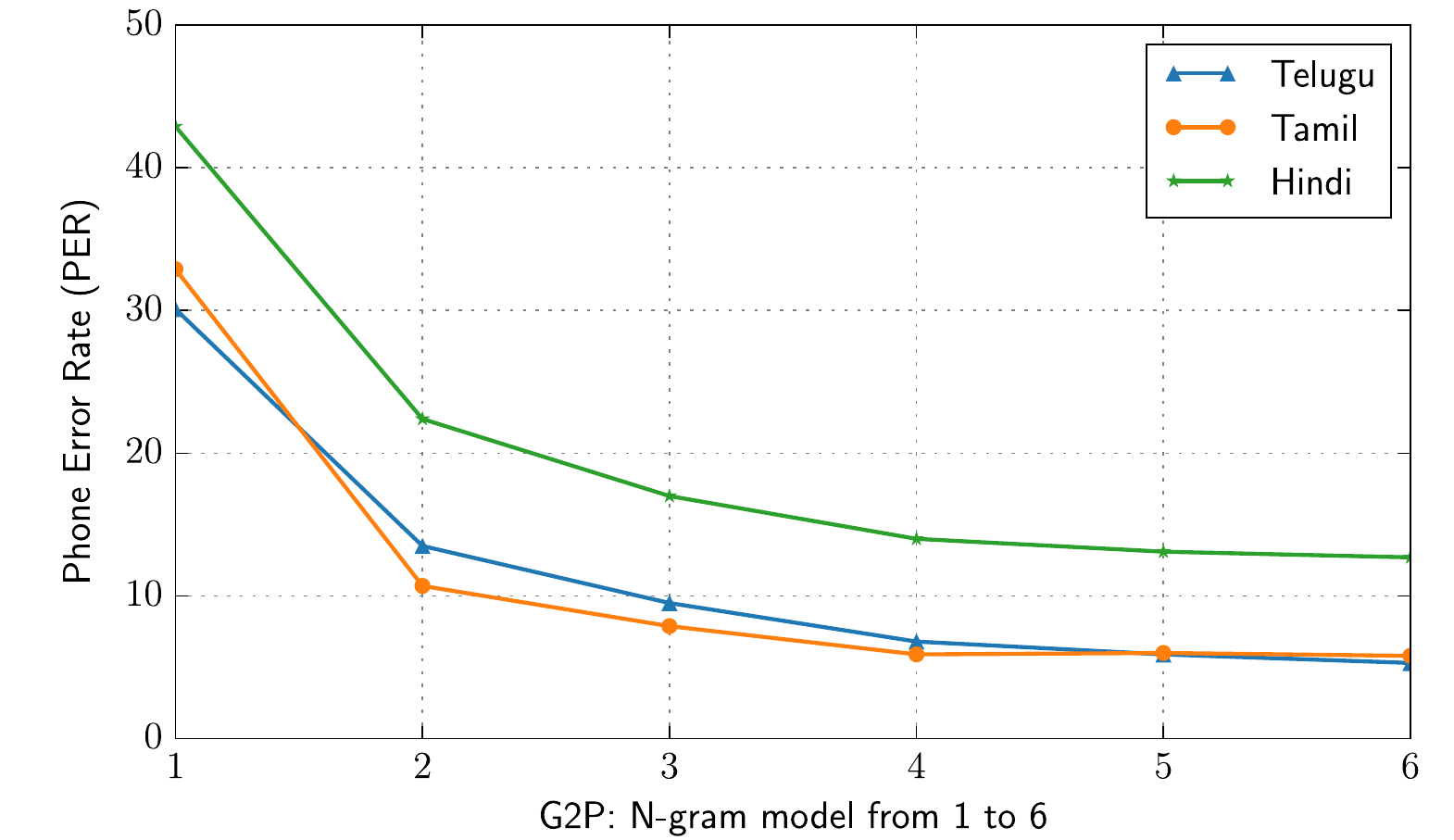}}
\caption{Performance of G2P models from uni-gram to 6-gram for Hindi, Telugu and Tamil}
\label{fig:g2p_performance}
\end{minipage}
\end{figure}

Given the pronunciation lexicon, we train a G2P transducer \cite{Bisani2008434} for each language separately with varying n-gram sequences. The corpus used for training is described in Section~\ref{ssec:data}. Figure \ref{fig:g2p_performance} displays the phone error rate of the G2P model with varying n-grams. The 6-gram model achieved the lowest phone error rate across the three languages. Telugu and Tamil achieved lower phonetic error rates compared to Hindi. This can be attributed to the ineffective handling of intricate \emph{schwa deletion}, a well-known phenomenon in Indo-Aryan languages. 

An advantage with this model is that, since the phoneme-set is standard, we can train G2P and DNN on two independent datasets -- G2P on  parallel transliterations of a very large corpus that could be obtained via crowdsourcing, and DNN model on gold phonetic speech transcriptions independently of the G2P model's performance. We leave this aspect of our work for future. In this work, we train a DNN model on the output of G2P aligned with natural speech.

\ignore{The phone error rate of Hindi is slightly higher when compared to Telugu and Tamil. This is partly because of the schwa deletion in manual transliteration and incorrect alignments during training. We also observed that some of the errors in G2P are substitutions are of short to long vowels,  /t/ for /th/, /d/ for /dh/, /s/ for /sh/ and vice-versa.  }

\vspace{-2mm}

\subsection{DNN Speech Synthesizer} 
\label{sec:approachesB}
We use a DNN for learning to synthesize speech from the phonetic strings obtained in the previous step. We use two independent DNNs -- one for duration and the other for acoustic modeling.

Let $x_{i}$ $=$ $[x_{i}(1),... ,x_{i}(d_{x})]^{T}$ and 
$y_{i}$ $=$ $[y_{i}(1),... ,y_{i}(d_{y})]^{T}$ be static input and output feature vectors of the DNN, where $d_{x}$ and $d_{y}$ denote the dimensions of $x_{i}$ and $y_{i}$, respectively, and $T$ denotes transposition. 

\textbf{Duration Model:} For duration modeling, the input comprises binary features ($x_p$) derived from a subset of the questions used by the decision-tree clustering in the standard HTS synthesiser. Similar to \cite{zen2013dnn, wu2015dnn}, frame-aligned data for DNN training is created by forced alignment using the HMM system. The output is an eight-dimensional vector ($y_p$) of durations for every phone, comprising five sub-state durations, the overall phone duration, syllable duration and whole word duration. We use this form of multi-task learning to improve the model; the three additional features (phone, syllable, and word durations) act as a secondary task to help the network learn more about suprasegmental variations in duration at word level. At synthesis time, these features are predicted, but ignored. 

\textbf{Acoustic Model:} The input uses the same features as duration prediction, to which $9$~numerical features are appended. These capture frame position in the HMM state and phoneme, state position in phoneme, and state and phoneme duration. The DNN outputs comprise MCCs, BAPs and continuous $logf_0$ (all with deltas and delta-deltas) plus a voiced/unvoiced binary value.  

In both acoustic and duration model, all the input features are normalized to the range of $[0.01, 0.99]$ and output features are normalized to zero mean and unit variance. The DNNs are then trained to map the linguistic features of input text to duration and acoustic  features respectively. If $D(x_i)$ denotes the DNN mapping of $x_i$, then the error of the mapping is given by:

\begin{equation}
  \epsilon = \sum ||y_i-D(x_i)||^{2}
\end{equation}
\begin{equation}
  D(x_i) = \widetilde{d}(z_{n+1})
\end{equation}
\begin{equation}
  z_{n+1}=d(w^{(n)}d(z_{n}))
\end{equation}
\begin{equation}
  d(\vartheta )=a \tanh (b\vartheta ), \widetilde{d}(\vartheta )=\vartheta
\end{equation}

\noindent where $n$ indexes layer and $w^{(n)}$ is the weight matrix of the $n^{th}$ layer of the DNN model. 

At synthesis time, duration is predicted first, and is used as an input to the acoustic model to predict the speech parameters. Maximum likelihood parameter generation (MLPG) using pre-computed variances from the training data is applied to the output features for synthesis, and spectral enhancement post-filtering is applied to the resulting MCC trajectories. Finally, the STRAIGHT vocoder \cite{bib5} is used to synthesize the waveform.

\section{Experimental Setup}
\label{sec:experiments}

\subsection{Speech Databases}
\label{ssec:data}

Our languages of interest are Hindi, Tamil and Telugu, all of which are widely-spoken Indian languages. We train and test on the 2015 Blizzard Challenge data which contains about four hours of speech and corresponding text for each language. The data-set contains 1710 utterances for Hindi, 1462 utterances for Tamil, and 2481 utterances for Telugu, with a single speaker per language.  We used 92\% of the data for training, 4\% for development and 4\% for testing. 

\subsection{Annotation}
Starting from the original transcriptions in native script, we asked crowdsourced human annotators to ASCII transliterate them using pronunciation as their main motivation for spelling. For Hindi and Tamil, we recruited paid workers via Mechanical Turk who could read and speak the language fluently (as self-reported); for Telugu we had access to a trusted pool of native speakers. 
We tokenize each sentence to words with whitespace and punctuations as the delimiters. 
The annotators were provided with a web-interface containing a text box for each word. This ensures transliteration of every word given in the input sentence. The total number of annotators for Telugu, Tamil and Hindi are 50, 66 and 82 respectively.
We diversified train, dev and test splits by having different set
of annotators for each split.

\subsection{Experimental Settings}
We used the same DNN architectures (Section \ref{sec:approachesB}) for both duration and acoustic modeling. The number of hidden layers used was 6 with each layer consisting of 1024 nodes. As shown in equation 4, the tanh function was used as the hidden activation function, and a linear activation function was employed at the output layer. During training, L2 regularization was applied to the weights with penalty factor of 0.00001, the mini-batch size was 256 for the acoustic model and 64 for the duration model. For the first 10 epochs, momentum was 0.3 with a fixed learning rate of 0.002. After 10 epochs, the momentum was increased to 0.9 and from that point on, the learning rate was halved at each epoch. The learning rate of the top two layers was always half that of other layers. Learning rate was fine-tuned in duration models to achieve best performance. The maximum number of epochs was set to 30 (i.e., early stopping). 

\subsection{Our Models}
As outlined in Section \ref{sec:approachesA}, we train three different models for each language. The number of questions used in DNN were different from system to system. For Uni-Grapheme model (labelled as \textbf{UGM}), the questions based on quin-phone identity were used, and other questions include suprasegmental features such as syllable, word, phrase and positional features. For Multi-Grapheme model (labelled as \textbf{MGM}) and Grapheme-to-Phoneme model (labelled as \textbf{G2P}), other questions based on position and manner of articulation were additionaly included. 

\subsection{Benchmark}
As a benchmark, we use the DNN speech synthesizer trained on CPS phonetic transcriptions of the speech data. The goal is thus to synthesize speech from ASCII text that is as close as possible in quality to this benchmark (labelled as \textbf{BMK}).

\section{Results}
\subsection{Objective Evaluations}

\begin{table*}[t]
\vspace{-0.5cm}
  \centering
    \caption{Objective results of the proposed techniques versus the benchmark approach. MCD and BAP are Mel-Cepstral Distortion and Band Aperiodicity distortion, respectively. V/UV error means frame-level voiced/unvoiced prediction error. Root Mean Squared Error (RMSE) of F0 was calculated on a linear frequency scale.}
  \begin{tabular}{|M{3.2cm}|M{1cm}M{1cm}M{1.5cm}M{1.5cm}|M{1cm}M{1cm}M{1.5cm}M{1.5cm}|}
    \hline
{\bf }       & {\bf \begin{tabular}[c]{@{}c@{}}MCD\\ (dB)\end{tabular}} & {\bf \begin{tabular}[c]{@{}c@{}}BAP\\ (dB)\end{tabular}} & {\bf \begin{tabular}[c]{@{}c@{}}F0\\ RMSE (Hz)\end{tabular}} & {\bf \begin{tabular}[c]{@{}c@{}}V/UV error\\ rate(\%)\end{tabular}} & {\bf \begin{tabular}[c]{@{}c@{}}MCD\\ (dB)\end{tabular}} & {\bf \begin{tabular}[c]{@{}c@{}}BAP\\ (dB)\end{tabular}} & {\bf \begin{tabular}[c]{@{}c@{}}F0\\ RMSE (Hz)\end{tabular}} & {\bf \begin{tabular}[c]{@{}c@{}}V/UV error\\ rate(\%)\end{tabular}} \\
        \hline
    {\bf Method} & \multicolumn{4}{c|}{{\bf Telugu}} & \multicolumn{4}{c|}{{\bf Hindi}}  \\ 
    \hline
  \end{tabular}
  \begin{tabular}{|M{3.2cm}|M{1cm}M{1cm}M{1.5cm}M{1.5cm}|M{1cm}M{1cm}M{1.5cm}M{1.5cm}|}
    \hline
    Uni-Grapheme-TTS & 4.97 & 2.05 & 35.13 & 6.23 & 4.82 & 1.92 & 10.87 & 9.25 \\
    Multi-Grapheme-TTS & 5.02 & 2.06 & 36.49 & 6.01 & 4.81 & 1.92 & 11.29 & 9.31 \\
    G2P-TTS & 4.77 & 2.04 & 35.18 & 5.94 & 4.80 & 1.92 & 10.77 & 9.30 \\
    Benchmark & 4.81 & 2.04 & 37.09 & 5.83 & 4.52 & 1.90 & 10.86 & 8.07 \\
    \hline
    {\bf Method} & \multicolumn{4}{c|}{{\bf Tamil}} & \multicolumn{4}{c|}{{\bf Combined Average}}  \\ 
    \hline
  \end{tabular}
  \begin{tabular}{|M{3.2cm}|M{1cm}M{1cm}M{1.5cm}M{1.5cm}|M{1cm}M{1cm}M{1.5cm}M{1.5cm}|}
    \hline
    Uni-Grapheme-TTS & 4.87 & 2.07 & 40.65 & 9.63 & 4.89 & 2.01 & 28.88 & 8.37 \\
    Multi-Grapheme-TTS & 5.15 & 2.09 & 43.31 & 9.92 & 4.99 & 2.02 & 30.36 & 8.41 \\
    G2P-TTS & 5.16 & 2.09 & 44.27 & 10.38 & 4.91 & 2.02 & 30.07 & 8.54 \\
    Benchmark & 5.06 & 2.08 & 43.67 & 10.04 & 4.79 & 2.01 & 30.54 & 7.98 \\
    \hline
  \end{tabular}
  \label{tab:objective}
\end{table*}

\begin{table}[t]
\vspace{-1.5em} 
\caption{RMSE (frames per phone) between predicted and forced-aligned durations. }
\begin{tabular}{|M{2cm}|M{1.3cm}M{1.3cm}M{1.3cm}|}
    \hline
    Models & Telugu & Hindi & Tamil  \\
    \hline
    UGM & 5.121 & 8.924 & 12.540 \\
    MGM & 5.015 & 9.876 & 13.105 \\
    G2P & 4.897 & 9.657 & 13.026  \\
    BMK & 4.118 & 9.321 & 12.378 \\
    \hline
  \end{tabular}
  \label{tab:dur_rmse}
  \end{table}
  
  \begin{table}[t]
  \vspace{-1.5em} 
  \caption{Pearson correlation between predicted and forced-aligned durations. }
\begin{tabular}{|M{2cm}|M{1.3cm}M{1.3cm}M{1.3cm}|}
    \hline
    Models & Telugu & Hindi & Tamil \\
    \hline
    UGM & 0.787 & 0.525 & 0.618 \\
    MGM & 0.807 & 0.533 & 0.624 \\
    G2P & 0.818 & 0.564 & 0.657 \\
    BMK & 0.866 & 0.692 & 0.695 \\
    \hline
  \end{tabular}
  \label{tab:dur_corr}
  \end{table}

\subsubsection{Duration Model}
To evaluate the duration prediction DNN, we calculated the root-mean-square error (RMSE) and Pearson correlation between reference and predicted durations, where the reference durations are estimated from the forced-alignment step in HTS. 
Tables \ref{tab:dur_rmse} and \ref{tab:dur_corr} present the results on test data. 

Overall, the benchmark system showed better performance than other systems in all languages. Among the proposed approaches, G2P performed slightly better than the other two in terms of correlation, whereas RMSE performance was not consistent across the languages. A possible explanation for this is that G2P uses superior phone set defined manually whereas UGM and MGM use unsupervised phones. Nevertheless, the proposed systems are not too far from the benchmark.

Compared to Telugu, Hindi and Tamil show worse objective scores. For these two languages, punctuation marks were not retained in the corpus, which made pauses harder to predict. As a consequence, occasional pauses in the acoustics were frequently forced-aligned to non-pause phones, introducing errors in the reference durations. These unpredictable elongations inflated the objective measures, without perturbing the actual predictions too much. (Telugu, in contrast, used oracle pauses, inserted using Festvox's \texttt{ehmm} based on the acoustics.)
  

\subsubsection{Acoustic Model}
 We used following four objective evaluations to assess the performance of the 
 proposed methods in comparison to the benchmark system.
 \begin{itemize}
     \item \textbf{MCD:} Mel-Cepstral Distortion (MCD) to measure MCC
prediction performance.
     \item \textbf{BAP:} to measure distortion of BAPs
     \item \textbf{F0 RMSE:} Root Mean Squared Error (RMSE) to measure the accuracy of F0 prediction. The error value was calculated on a linear scale instead of log-scale which was used to model the F0 values.
     \item \textbf{V/UV:} to measure voiced/unvoiced error.
 \end{itemize}
In all these metrics, a lower value gives the better performance. While the objective metrics do not map directly to perceptual quality, they are often useful for system tuning.
Table \ref{tab:objective} presents the results on test data.
As expected, the benchmark model performs well on most metrics. While the G2P Model performs well on Telugu and Hindi, the Uni-Grapheme model does well on Tamil. Overall, the proposed approaches compare favourably with the benchmark.

\subsection{Subjective Evaluations}

Three MUSHRA (MUltiple Stimuli with Hidden Reference and Anchor)\footnote{\url{https://github.com/HSU-ANT/beaqlejs}} \cite{beaqlejs2014html} tests were conducted to assess the naturalness of the synthesized speech. For each language, $16$ native listeners were exposed to 20 sentences, chosen randomly from the test set. For each sentence, $5$ unlabelled stimuli were presented in parallel: one for each of the four synthesis systems speaking that sentence plus copy-synthesis speech (i.e., vocoded speech, labelled as \textbf{VOC}) used as the hidden reference. Listeners were asked to rate each stimulus from $0$ (extremely bad for naturalness) to $100$ (same naturalness as the reference speech), and also instructed to give exactly one of the $5$ stimuli in every set a rating of $100$.

\begin{figure}[t]
\begin{minipage}[b]{1.0\linewidth}
  \centering
  \centerline{\includegraphics[width=8.5cm]{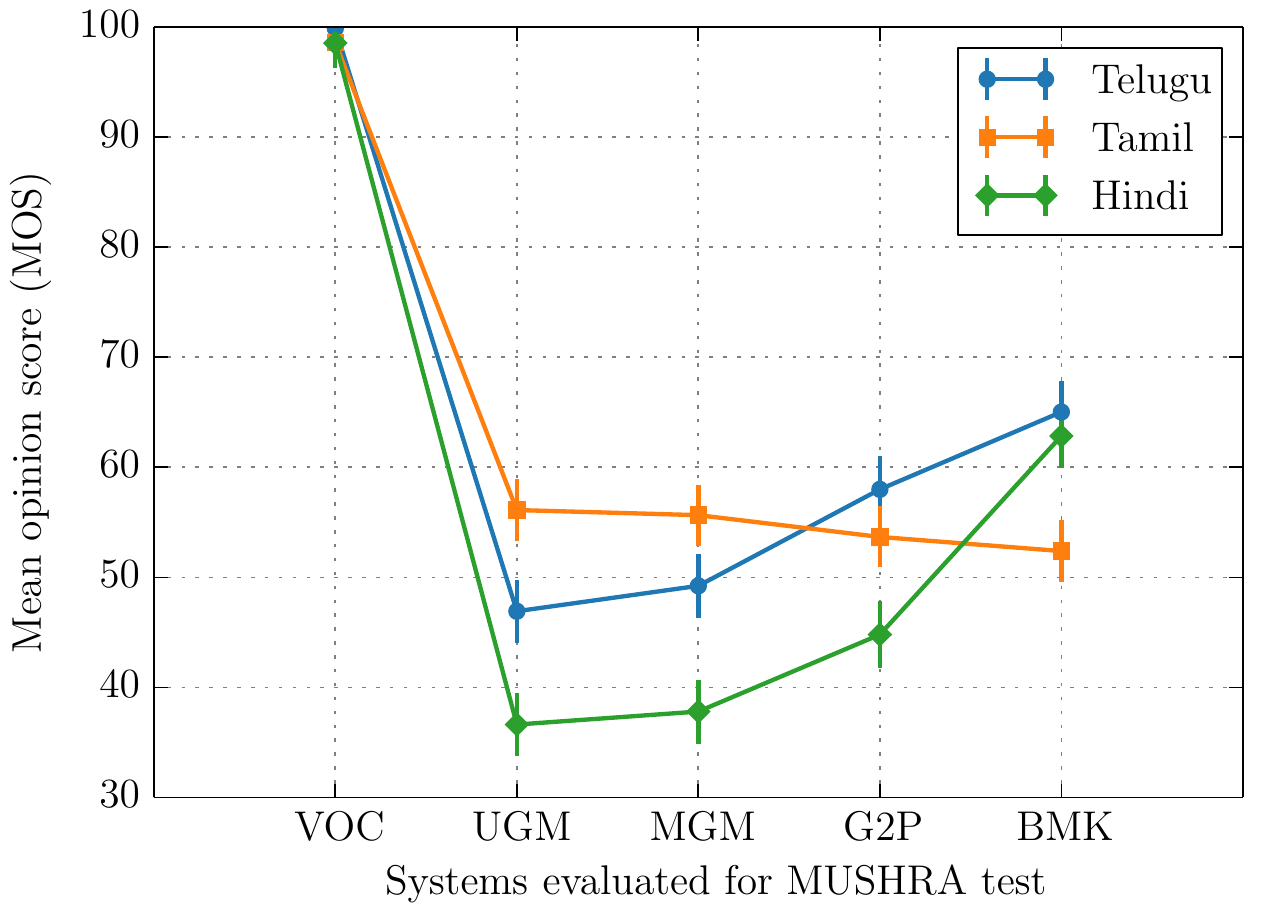}}
\caption{Performance of systems evaluated in the MUSHRA test for all three languages.}
\label{fig:MUSHRA_MOS}
\end{minipage}
\vspace{-0.5cm}
\end{figure}

\begin{figure}[h]
\begin{minipage}[b]{1.0\linewidth}
  \centering
  \centerline{\includegraphics[width=8.5cm]{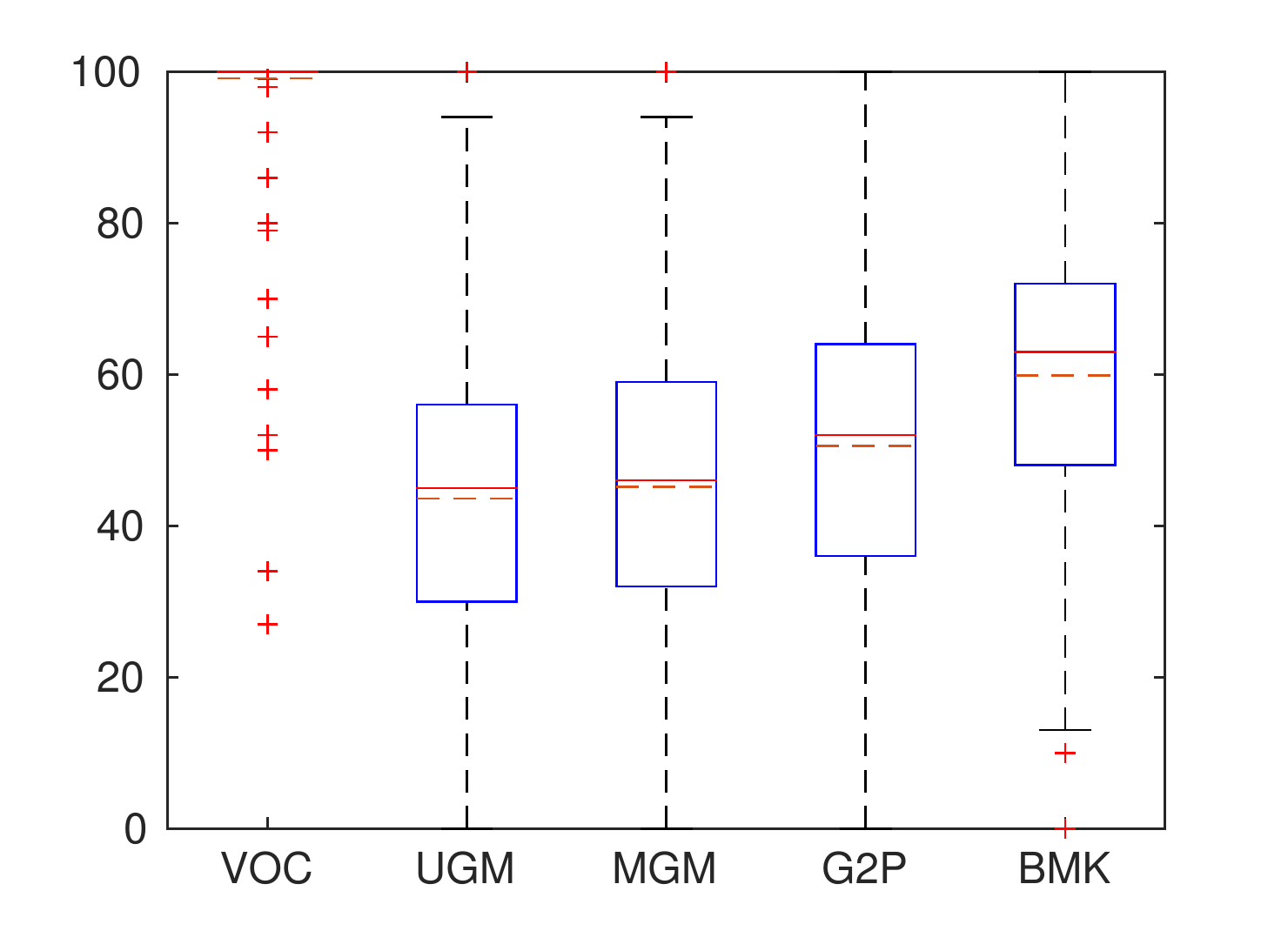}}
\caption{Box plot of absolute values from all three languages' listening tests. Red lines are medians, dashed lines means. Box edges show quartiles. Plus signs indicate outliers.}
\label{fig:new_results_mos}
\end{minipage}
\vspace{-0.5cm}
\end{figure}

\begin{figure}[t]
\begin{minipage}[b]{1.0\linewidth}
  \centering
  \centerline{\includegraphics[width=8.5cm]{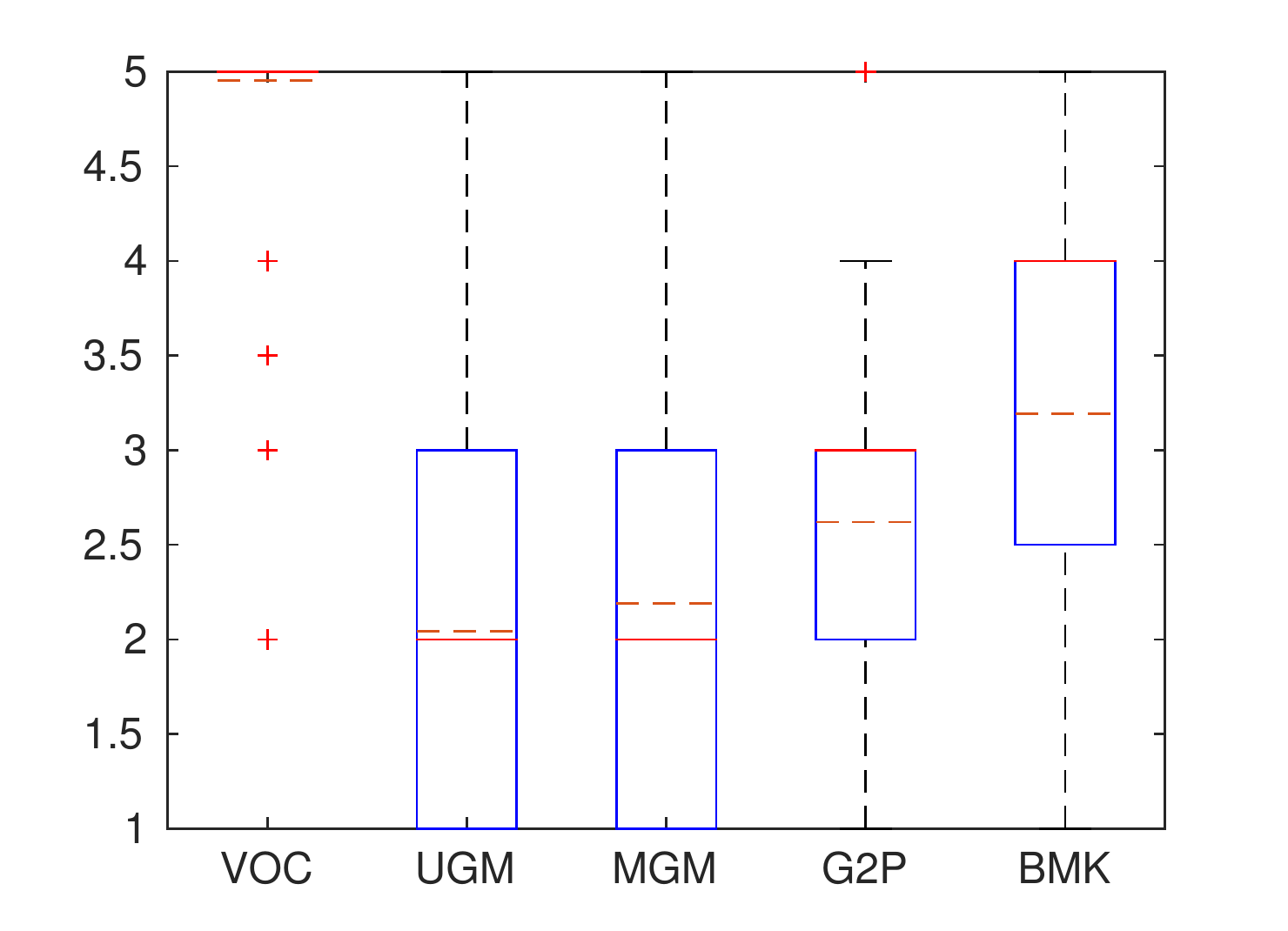}}
\caption{Box plot of aggregate ranks from listening tests (higher is better). Red lines are medians, dashed lines means. Box edges show quartiles. Plus signs indicate outliers.}
\label{fig:new_results_ranks}
\end{minipage}
\vspace{-0.5cm}
\end{figure}

\begin{figure}[h]
\begin{minipage}[b]{1.0\linewidth}
  \centering
  \centerline{\includegraphics[width=8.5cm]{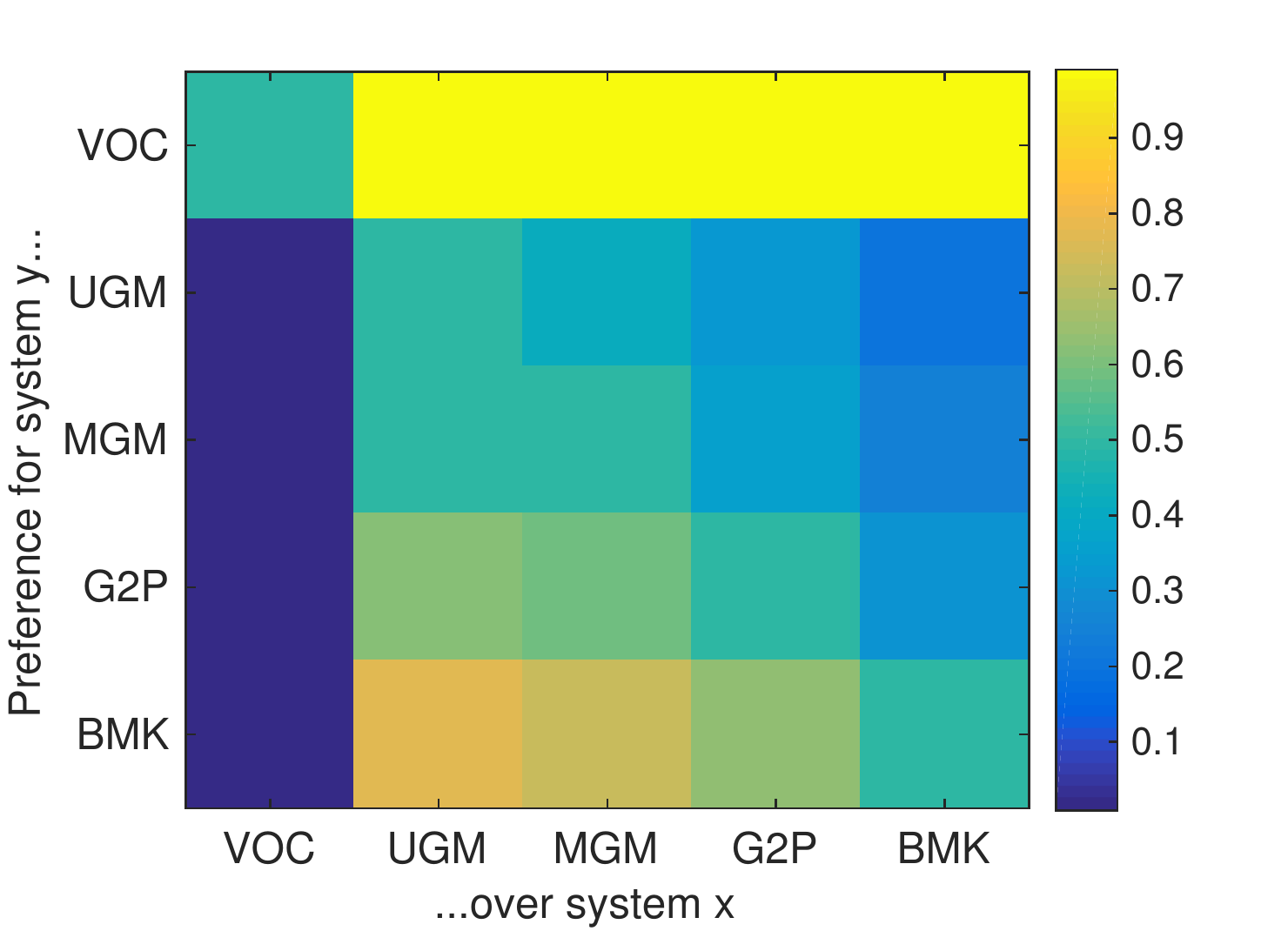}}
\caption{Preferences among systems (how often $y$ was rated above $x$).}
\label{fig:preference_over_systems}
\end{minipage}
\vspace{-0.5cm}
\end{figure}

For Telugu and Hindi, we had access to a trusted pool of native speakers from IIIT-Hyderabad, while for Tamil we recruited paid workers via Amazon Mechanical Turk as listeners. The Mean Opinion Scores (MOS) from the tests are presented in Figure \ref{fig:MUSHRA_MOS} with their standard deviation represented in log-scale. The benchmark model achieves a higher MOS in Telugu and Hindi, as expected, while in Tamil the Uni-Grapheme model achieves best performance. However, according to paired $t$-tests with Holm-Bonferroni correction for multiple comparisons, the difference with next best system is significant only in Telugu and Hindi.
Among the proposed approaches, G2P performed significantly better than other two in Telugu and Hindi. However in Tamil, both G2P and benchmark performed worse than the rest. This strange behaviour can be attributed to two reasons: 1) the absence of a mechanism for detecting outliers in turker judgements (as opposed to the use of trusted pool of listeners for Hindi and Telugu); 2) the lack of our expertize in enhancing letter to sound rules specific to Tamil. The difference in ratings suggest that some additional rules or fine-tuning of lexicon may be required for Tamil.

The MUSHRA scores combined across all three languages for each system are presented in Fig. \ref{fig:new_results_mos}. For further analysis, each set of fifteen parallel listener scores was converted
to ranks from 1 (worst) to 5 (best), with tied ranks set
to the mean of the tied position. A box plot of these rank scores
aggregated across all sentences and listeners is shown in Figure \ref{fig:new_results_ranks}. Listener preferences between systems are also illustrated in Figure \ref{fig:preference_over_systems}. All these figures indicate, G2P performed the best among the proposed approaches. 

An interesting issue is that some test sentences include English-language words (e.g.: road, page, congress) due to frequent code-switching among the native speakers (also reflected in the text corpus). This affected the performance of G2P conversion for those sentences, in turn creating a marginal difference between G2P and benchmark over the listening test. G2P trained on large corpora of parallel text may remove such errors in the future, thereby improving the synthesis quality and reducing the gap towards the benchmark. \cite{sunanya2016codeswitch} is one such recent attempt for synthesizing speech from code-mixed text. 

No intelligibility evaluation was conducted since transcription word error rate (WER) has been found to be a poor metric for Indian languages, cf.\ \cite{prahallad2014blizzard}. However, we believe listeners do take into account intelligibility while rating the stimuli, even though they were asked to rate the naturalness.

\section{Applications}
The grapheme-to-phoneme conversion described herein enabled us to build indic-search\footnote{\url{http://srikanthr.in/indic-search}}, a search engine that helps end-users use ASCII to search for pages written in Unicode.
Text-to-speech interfaces with ASCII input also enable users to type in their own pronunciation rather than conforming to a specific notation.

\section{Conclusions}
\label{sec:print}

In this paper, we considered the problem of synthesizing speech from ASCII transliterated text of Indian languages. Our proposed approach first converts ASCII text to phonetic script, and then learns a DNN to synthesize speech from the phonetic script. We experimented with three approaches, which vary in the degree of manual supervision in defining phonemes. Our results show that G2P model with few assumptions is competitive with manually-defined phoneme models. All the data, and samples used in the listening tests are available online at: \footnotesize{\url{http://srikanthr.in/indic-speech-synthesis}}.

\newpage
 \textbf{Acknowledgements: }Thanks to Nivedita Chennupati and Spandana Gella for their contribution in data collection with Amazon Mechanical Turk. Also, thanks to Sivanada Achanta for evaluating the systems through listening tests. We thank Gustav Henter for proofreading. However, the errors that remain are the authors' responsibilities.

  \bibliographystyle{IEEEtran}

  \bibliography{mybib}


\end{document}